\documentclass{article}

\usepackage{ijcai16}

\usepackage{times}
\usepackage{epsfig}
\usepackage{graphicx}
\usepackage{amsmath}
\usepackage{amssymb}
\usepackage{mathsymb}
\usepackage{textcomp}
\usepackage[linesnumbered,algoruled,boxed,lined]{algorithm2e}
\usepackage{verbatim}
\usepackage{subfigure}
\usepackage{url}
\usepackage{graphicx}
\usepackage{multirow}
\usepackage[super]{nth}
\usepackage{balance}
\usepackage{colortbl}
\graphicspath{{./}{./Fig/}{./Figs/}{./Fig/eps/}{./Fig/pdf/}}

\usepackage{algorithm2e}

\let\savedalgorithm\algorithm
\let\savedendalgorithm\endalgorithm

\usepackage{algorithm}

\newcommand{\etal}{\textit{et al.\ }}
\newcommand{\eg}{\emph{e.g.,\ }}
\newcommand{\ie}{\emph{i.e.,\ }}

\title{Robust Hashing for Multi-View Data: Jointly Learning Low-Rank Kernelized Similarity Consensus and Hash Functions}
\author{Lin Wu$^{\ddag}$, Yang Wang$^{\dag}$\\
 $^{\ddag}$The University of Adelaide, Australia\\
 $^{\dag}$The University of New South Wales, Kensington, Sydney, Australia\\
 lin.wu@adelaide.edu.au \space\space wangy@cse.unsw.edu.au
}

\begin{document}

\maketitle

\begin{abstract}

Learning hash functions/codes for similarity search over multi-view data is attracting increasing attention, where similar hash codes are assigned to the data objects characterizing consistently neighborhood relationship across views. Traditional methods in this category inherently suffer three limitations:  1) they commonly adopt a two-stage scheme where similarity matrix is first constructed, followed by a subsequent hash function learning; 2) these methods are commonly developed on the assumption that data samples with multiple representations are noise-free,which is not practical in real-life applications; 3) they often incur cumbersome training model caused by the neighborhood graph construction using all $N$ points in the database ($O(N)$).
In this paper, we motivate the problem of jointly and efficiently training the robust hash functions over data objects with multi-feature representations which may be noise corrupted. To achieve both the robustness and training efficiency, we propose an approach to effectively and efficiently learning low-rank kernelized \footnote{We use kernelized similarity rather than kernel, as it is not a squared symmetric matrix for data-landmark affinity matrix.} hash functions shared across views. Specifically, we utilize landmark graphs to construct tractable similarity matrices in multi-views to automatically discover neighborhood structure in the data. To learn robust hash functions, a latent low-rank kernel function is used to construct hash functions in order to accommodate linearly inseparable data. In particular, a latent kernelized similarity matrix is recovered by rank minimization on multiple kernel-based similarity matrices. Extensive experiments on real-world multi-view datasets validate the efficacy of our method in the presence of error corruptions.

\end{abstract}

\section{Introduction}\label{sec:intro}

Hashing is dramatically efficient for similarity search over low-dimensional binary codes with low storage cost. Intensive hashing methods valid on single data source have been proposed which can be classified into \emph{data-independent} hashing such as locality sensitive hashing (LSH) \cite{socg04} and \emph{data-dependent} hashing or learning based hashing \cite{Spectral-hash,PCAH}.

In real-life situations, data objects can be decomposed of multi-view (feature) spaces where each view can characterize its individual property, \eg an image can be described by color histograms and textures, and the two features turn out to be complementary to each other \cite{WangMM14,WangCIKM,WangSIGIR15,LinMM13,LinTcyb2016,WangIJCAI16,WangTIP2015,WangMetricFusion2015,WangKAIS}. Consequently, a wealth of multi-view hashing methods \cite{Composite-hash,ECCV12,MSPH,MM11,MM12,MVLH} are developed in order to effectively leverage complementary priors from multi-views to achieve performance improvement in similarity search. The critical issue is to ensure the learned hash codes can well preserve the original data similarities regarding view-dependent feature representations. To be specific, similar hash codes are assigned to data objects that consistently capture nearest neighborhood structure across all views.

\subsection{Motivation}
Despite improved performance delivered by existing multi-view hashing methods \cite{Composite-hash,ECCV12,MSPH,MM11,MM12,MVLH}, some fundamental limitations can be identified:

\begin{itemize}
\item The learning process is conducted by a two-stage mechanism where hash functions are learned based on pre-constructed data similarity matrix. Their methods commonly assume that data samples are noise-free under multiple views whereas in real-world applications input data objects may be noisy (\eg missing values in pixels), resulting in corresponding similarity matrices being corrupted by considerable noises \cite{WangTIP2015,RobustMVSC}.
Moreover, the recovery of \emph{consensus} or \emph{requisite} similarity values across views in the presence of noise contamination remains an unresolved challenge  in multi-view data analysis \cite{MVLowRank,MVLow-rank-regression,RobustLateFusion}.

This motivates us to deliver a framework to jointly and effectively learn similarity matrices and \emph{robust} hash functions with kernel functions plugged because the kernel trick is able to tackle linearly inseparable data \cite{Supervised-hash-kernel}. To this end, a \emph{latent}  kernelized similarity matrix is recovered shared across views by using low-rank representation (LRR) \cite{LiuLRR-PAMI13} which is robust to corrupted observations. The recovered \emph{low-rank} kernelized similarity matrix is consensus-reaching across views and can reveal the true underlying structures in data points.

\item State-of-the-art multi-view hashing methods is less efficiency in their learning procedure because the learning is performed by building and accessing a neighborhood graph using all $N$ points ($O(N^2)$). This action is intractable in off-line training when $N$ is large.

To this end, we are further motivated to employ an \emph{landmark graph} to build an approximate neighborhood graph using landmarks \cite{Hash:graph,Liu:anchor2010}, in which the similarity between a pair of data points is measured with respect to a small number of landmarks (typically a few hundred). The resulting graph is built in $O(N)$ time and sufficiently sparse with performance approaching to true $k$-NN graphs as the number of landmarks increases \cite{Hash:graph}.

\end{itemize}

\subsection{Our Method}
In this paper, we propose a novel approach to robust multi-view hashing by effectively and efficiently learning a set of hash functions and a low-rank kernelized similarity matrix shared by multiple views.

We remark that our method is fundamentally different from existing multi-view hashing methods that are conditioned on corruption-free similarities, which has diminished their application to real-world tasks. Instead, we propose to learn hash functions and kernel-based similarities under a more realistic scenario with noisy observations. Our method is advantageous in the aspect of efficiency due to the employment of approximate neighborhood with landmark graphs.
We clarify the recovered low-rank similarity matrix in kernel functions to be the kernelized rather than kernel since it is not a symmetric matrix yet characterizes non-linear similarities. The proposed method is also different from partial view study \cite{Co-training,HashPartial}, where they consider the case that data examples with some modalities are missing. Our approach follows the setting of multi-view learning which aims to improve existing single view model by learning a model utilizing data collected from multiple channels \cite{MVSurvey,MVLow-rank-regression,NIPS11,WangMM14,WangMM15,RobustMVSC} where all data samples have full information in all views.

In our framework, the low rank minimization is enforced to yield a consensus-reaching, kernelized similarity matrix shared by multiple views where larger similarity values indicate corresponding data objects from the same cluster, while smaller similarity values imply those come from distinct clusters. Thus, the learned low-rank similarity matrix against multi-views can reflect the underlying clustering information.

Technically, a nonlinear kernelized similarity matrix in the $m$-th view, denoted as $K^{(m)}$, can be decomposed into three components: (1) A latent low-rank kernelized similarity matrix $\hat{K}$, representing the nonlinear requisite or consensus similarities shared across views; (2) a view-dependent redundancy characterizing its individual similarities; and (3) possible error corruptions for view-specific representations. We unify view redundancy and errors into $E^{(m)}$ and impose an $\ell_{2,1}$-norm constraint on it, denoted as  $||E^{(m)}||_{2,1}$. This is because view redundancy and disturbing errors are always sparsely distributed, and minimizing $||E^{(m)}||_{2,1}$ is able to identify non-zero sparse columns revealing corresponding redundancy/errors.
Note that in this work, ``error" generally refers to error corruptions or perturbation, \eg noise or missing values, in view-dependent feature values. These principles are formulated into an objective function, which is optimized based on the inexact Augmented Lagrangian Multiplier (ALM) scheme \cite{ALM-report}. It allows us to jointly learn a latent low-rank nonlinear similarity with corruption free and optimal hash functions for multi-view data, where hash codes are restricted to well preserve local (neighborhood) geometric structures in each view. We remark that several cross-view semantic hashing algorithms \cite{example3,example4,IJCAICross} have been developed to embed multiple high dimensional features from \emph{heterogeneous} data sources into one Hamming space, while preserving their original similarities.
Our setting is fundamentally different from cross-view/modal hashing in the aspect that we aim to leverage multiple features to jointly learn hash functions and a latent nonlinear similarity matrix over a \emph{homogeneous} data source. To the best of our knowledge, we are the first to systematically address the problem of multi-view hashing with possible data error corruptions.

\subsection{Contributions}
The major contributions of this paper are three-fold.

\begin{itemize}
\item We motivate the problem of robust hashing over multi-view
data with nonlinear data distribution, and propose to learn the robust hash functions and a low-rank kernelized similarity matrix shared by views.

\item An iterative low-rank recovery optimization technique is proposed to learn the robust hashing functions. For the sake of efficiency, the neighborhood graph is approximated by using landmark graphs with sparse connection between data points.

\item Extensive experiments conducted on real-world multi-view datasets validate the efficacy of our method in the presence of error corruptions for multi-view feature representations. 
\end{itemize}



\section{Related Work}\label{sec:related}

\subsection{Multi-view Learning based Hashing}
The purpose of multi-view learning based hashing is to learn better hash codes by leveraging multiple views. Some recent representative works include Multiple Feature Hashing (MFH) \cite{MM11}, Composite Hashing with Multiple Sources (CHMS) \cite{Composite-hash}, Compact Kernel Hashing with multiple features (CKH) \cite{MM12}, and Multi-view Sequential Spectral Hashing (SSH) \cite{ECCV12}. However, these methods have common drawbacks that they typically apply spectral graph technique (\eg $k$-NN graph) to model a similarities between data points. In general, the complexity of constructing the similarity matrix is $\mathcal{O}(N^2)$ for $N$ data points, which is not pragmatic in large-scale applications. Moreover, the similarity matrix induced by graph construction is very sensitive to noise corruptions. To avoid the construction of similarity matrix, Shen \etal \cite{MVLH} present a Multi-View Latent Hashing (MVLH) to learn hash codes by performing matrix factorization on a unified kernel feature space over multiple views.
Nonetheless, there are significant differences between MVLH and our approach. First, matrix factorization is performed on a unified kernel space which is formed by simply concatenating multiple kernel feature spaces. This would discard distinct local structures in individual views. By contrast,  the kernelized similarity matrix is constructed with respect to the distinct characteristic in each view.  Second, MVLH neglects the case of potential noise corruption in data samples. In this aspect, we attentively employ the low-rank representation (LRR) \cite{LiuLRR-PAMI13} to recover latent subspace structures from corrupted data.

\subsection{Low-rank Modeling}
Low-rank modeling in attracting increasing attention due to its capability of recovering the underlying structure among data objects \cite{RobustPCA,ExactMatrixCom,MVLowRank,ZhangJointLR,SimilarityLRR,BilinearLR}. It has striking success in many applications such as data compression \cite{RobustPCA}, subspace clustering \cite{LiuLRR-PAMI13,LRHeuristic,SimilarityLRR}, and image processing \cite{MovingLRR,ZhangJointLR,BilinearLR}. For instance, in \cite{ZhangJointLR}, Zhang \etal consider a joint formulation of recovering low-rank and sparse subspace structures for robust representation.

Nowadays, data are usually collected from diverse domains or obtained from various feature extractors, and each group of features can be regarded as a particular view \cite{MVSurvey}. Moreover, these data can be easily corrupted by potential noises (\eg missing pixels or outliers), or large variations (\eg post variations in face images) in real applications.
In practice, the underlying structure of data could be multiple subspaces, and thus Low-Rank Representation (LRR) is designed to find subspace structures in noisy data \cite{LiuLRR-PAMI13,SimilarityLRR}.
The multi-view low-rank analysis \cite{MVLowRank} is a recently proposed multi-view learning approach, which introduces low-rank constraint to reveal the intrinsic structure of data, and identifies outliers for the representation coefficients in low-rank matrix recovery.

In this paper, we are the first to apply low-rank learning to reveal structured kernalized similarity among multi-view data, and scale it up well to large-scale applications.

\section{Robust Multi-view Hashing}\label{sec:tech}

\subsection{Preliminary and Problem Definition}

Let $\phi(\cdot) = \{\phi_1(\cdot),\cdot\cdot,\phi_M(\cdot)\}$ be the embedding function for $M$ nonlinear feature spaces, each of which corresponds to one view. Following the Kernelized Locality Sensitive Hashing \cite{KLSH}, we uniformly select $R$ samples from the training set $X$, denoted by $Z_r$ ($r=1,\ldots,R$), to construct kernelized similarity matrices under multiple views. Given a sample represented by its feature $x_i$, the $p$-th hash bit can be generated via the linear projection:
\begin{equation}
h_p(x_i)=sign(C_p^{T} \phi(x_i) + b_p),
\end{equation}
where $sign(\cdot)$ denotes the element-wise function, which is 1 if it is larger or equal to 0 and -1 otherwise. $C_p = \sum_{r = 1}^{R} W_{rp} \phi(Z_r)$ indicates the linear combination of $R$ landmarks,
which can be the cluster centers \cite{Hash:graph} via scalable $R$-means clustering over the feature space with $d$ dimensions. $b_p \in \mathbb{R}$ is a bias term. Then, we have
\begin{equation}\label{eq:hashbit}
h_p(x_i) = sign(\sum_{r = 1}^{R}W_{rp}K_i + b_p),
\end{equation}
where $K_i$ denotes the $i$-th column of $K \in \mathbb{R}^{R \times N}$, such that $K = \sum_{i = 1}^{M}K^{(m)}$, and $K^{(m)} \in \mathbb{R}^{R\times N} (m=1,\ldots,M)$ denotes the kernelized similarity matrix between $R$ landmarks and $N$ samples corresponding to the kernelized  representation $\phi^{(m)}(\cdot)$. Accordingly, the hash code of $x_i$ can be rewritten via the kernel form,
\begin{equation}\label{eq:finalhashcode}
y_i=sign(W^T K_i + b),
\end{equation}
where $W \in \mathbb{R}^{R \times P}$ and $b = [b_1,\ldots, b_P]$.

Given a set of training samples $X=[x_1,\ldots,x_N]$ that may contain errors, $x_i^{(m)}\in \mathbb{R}^{d_m\times 1}$ denotes the $m$-th feature of $x_i$, and $d_m$ is the dimensionality for the feature space regarding the $m$-th view. Then $X^{(m)}=[x_1^{(m)},x_2^{(m)}\ldots,x_N^{(m)}] \in \mathbb{R}^{d_m \times N}$ is the view matrix corresponding to the $m^{th}$ feature of all training data. $x_i=[(x^1)^T_i,\ldots,(x^M)^T_i]^T \in \mathbb{R}^{d\times 1}$ is the vector representation of the $i^{th}$ training data using all features where $d=\sum_{m=1}^M d_m$, and $M$ is the number of views. We denote $Y=[y_1,y_2,\ldots,y_N] \in \mathbb{R}^{P \times N}$ as the hash codes of the training samples corresponding to all features, and $Y^{(m)}=[y_1^{(m)},\ldots,y_N^{(m)}] \in R^{P \times N}$ as the hash codes of the training data for the $m$-th view.
We aim to learn a latent low-rank kernel matrix $\hat{K}$ shared across multiple kernels, and construct a set of robust hashing functions $H = \{h_1(\cdot),\ldots,h_P(\cdot)\}$ for multi-view data where $h_p: \mathbb{R}^d \mapsto \{1,-1\}$ ($p=1,2,\ldots, P$), and $P$ is the number of hashing functions, \ie the hash code length.
The kernel function is plugged into hash function because the kernel trick has been theoretically and empirically proved to be able to tackle the data distribution that is almost linearly inseparable \cite{Supervised-hash-kernel}.

\subsection{Low-rank Kernelized Similarity Recovery from Multi-views}

Given a collection of high-dimensional multi-view data samples that may contain certain errors for each view-specific representation, we construct multiple nonlinear feature spaces $K^{(m)}(m=1,\ldots,M)$, each of which represents one feature view. To leverage multiple complementary representations, we propose to derive a consensus low-rank kernelized similarity matrix $\hat{K}$ recovered from corrupted data objects, and shared across views. This low-rank nonlinear similarity matrix is considered as the most requisite component, whilst each view also contains individual non-requisite information including redundancy and errors.
We explicitly model the redundancy via sparsity since multi-view study suggests that each individual view is sufficient to identify most of the similarity structure, and the deviation between requisite component and data sample is sparse \cite{NIPS11}.
In reality, data samples can be grossly corrupted due to the sensor failure or communication errors. Thus, an $\ell_{2,1}$-norm is adopted to characterize errors since they usually cause column sparsity in an affinity matrix \cite{LiuLRR-PAMI13}.

In our framework, the low-rank similarity matrix is constructed to be sparse by considering data samples and landmarks, thus ascertaining the efficiency of our approach.
Therefore, the latent low-rank kernelized similarity matrix $\hat{K}$ can be recovered from $K^{(m)} (m=1,\ldots,M)$ through a low-rank constraint on $\hat{K}$ and sparse constraint on each $E^{(m)}$, that is,
\begin{equation}\label{eq:consensus}
\min_{\hat{K},E^{(m)}} ||\hat{K}||_\ast + \lambda \sum_{m=1}^M ||E^{(m)}||_{2,1}, ~~ s.t.~~ K^{(m)}=\hat{K} + E^{(m)},  \hat{K} \geq 0.
\end{equation}
where $\lambda$ is the trade-off parameter and $E^{(m)}$ encodes the summation of error corruption and possible noise information regarding the $m$-th view.

\subsection{Objective Function}
Many studies \cite{Spectral-hash,MM11} have shown the benefits to exploit local structure of the training data to infer accurate and compact hash codes. However, all these algorithms are sensitive to error corruptions, hampering them to be effective in practical situations. By contrast, we propose to jointly learn hash codes by preserving local similarities in multiple views while being robust to errors. To exploit the local structure in each view, we define $M$ affinity matrices $S^{(m)}\in \mathbb{R}^{N \times N} (m=1,\ldots,M)$, one for each view, that is,
\begin{displaymath}\small
S^{(m)}_{ij}=\left\{
               \begin{array}{ll}
                 1, & \hbox{$x_i^{(m)} \in \mathcal{N}_k(x_j^{(m)})$ or $x_j^{(m)} \in \mathcal{N}_k(x_i^{(m)})$;} \\
                 0, & \hbox{else.}
               \end{array}
             \right.
\end{displaymath}
where $\mathcal{N}_k(\cdot)$ is the $k$-nearest neighbor set, and the Euclidean distance is employed in each feature space to determine the neighborhood.
A reasonable criteria of learning hash codes $y_i^{(m)}$ from the $m$-th view is to ensure similar objects in the original space should have similar binary hash codes. This can be formulated as below:
\begin{equation}
\min \sum_{i,j=1}^N S^{(m)}_{ij} ||y_i^{(m)}-y_j^{(m)}||_F^2.
\end{equation}

Given a training sample $x_i$, we expect the optimal hash code $y_i$ consistent with its distinct hash codes $y_i^{(m)}$ derived from each view. In this way, the local geometric structure in a single view can be globally optimized.
Therefore, we have
\begin{equation}
\min \sum_{m=1}^M \left(\sum_{i,j=1}^N S^{(m)}_{ij} ||y_i^{(m)}-y_j^{(m)}||_F^2 +\gamma \sum_{i=1}^N ||y_i-y_i^{(m)}||_F^2\right),
\end{equation}
where $\gamma$ is a trade-off parameter. The main bottleneck in the above formulation is computation where the cost of building the underlying graph and its associate affinity matrix $S^{(m)}$ is $O(d_m N^2)$, which is intractable for large $N$. To avoid the computational bottleneck, we employ a landmark graph by using a small set of $L$ points called landmarks to approximate the data neighborhood structure \cite{Hash:graph}. Similarities of all $N$ database points are measured with respect to these $L$ landmarks, and the true adjacency/similarity matrix $S^{(m)}$ in the $m$-th view is approximated using these similarities. First, K-means clustering \footnote{In practice, running K-means algorithm on a small subsample of the database with very few iterations is sufficient.} is performed on $N$ data points to obtain $L$ ($L \ll N$) clusters center $\mathcal{U}=\{u_j\in \mathbb{R}^{d_m}\}_{j=1}^L$ that act as landmark points. Next, the landmark graph defines the truncated similarities $F_{ij}$'s between all $N$ data points and $L$ landmarks as,
\[ F_{ij}=\left\{\begin{array}{cl}
\frac{\exp (-\mathcal{D}^2 (x_i,u_j)/t)}{\sum_{j'\in \langle i \rangle} \exp (-\mathcal{D}^2 (x_i,u_j')/t)},& \forall j\in \langle i \rangle \\
0,& \mbox{elsewhere}\end{array}\right.\]
where $\langle i \rangle \subset [1:L]$ denotes the indices of $k$ ($k\ll L$) nearest landmarks of points $x_i$ in $\mathcal{U}$ according to a distance function $\mathcal{D}()$ such as $\ell_2$ distance, and $t$ denotes the bandwidth parameter. Note that the matrix $F \in \mathbb{R}^{N\times L}$ is highly sparse. Each row of $F$ contains only $k$ non-zero entries which sum to 1. Thus, the landmark graph provides a powerful approximation to the adjacency matrix $S^{(m)}$ as $\hat{S}^{(m)}=F \Lambda^{-1} F^T$ where $\Lambda = \mbox{diag}(F^T \textbf{1}) \in \mathbb{R}^{L\times L}$ \cite{Hash:graph}.

For ease of representation, we denote $\mathcal{L}(Y,Y^{(m)})=\sum_{i,j=1}^N \hat{S}^{(m)}_{ij} ||y_i^{(m)}-y_j^{(m)}||_F^2 +\gamma \sum_{i=1}^N ||y_i-y_i^{(m)}||_F^2$.
To learn a set of hashing functions and a consensus nonlinear representation in a joint framework, we formulate the objective function of robust multi-view hashing as follows
\begin{equation}\label{eq:obj}
\begin{split}
&\min_{W,\hat{K},b,E^{(m)}} \sum_{m=1}^M \mathcal{L}(Y,Y^{(m)}) +\alpha ||\hat{K}||_\ast + \lambda \sum_{m=1}^M ||E^{(m)}||_{2,1}, \\
&s.t.~~ K^{(m)}=\hat{K} + E^{(m)}, \hat{K} \geq 0, m=1,\ldots,M, \\
&y_i=sign(W^T \hat{K}_i + b) \in \{-1,1\}^P, YY^T=I,
\end{split}
\end{equation}
where $\alpha$ is a trade-off parameter, $y_i \in \{-1,1\}^P$ enforces the hash code $y_i$ to be binary codes, and the constraint $YY^T=I$ is imposed to encourage bit de-correlations while avoiding the trivial solution. Due to the discrete constraints and non-convexity, the optimization problem in Eq.\eqref{eq:obj} is difficult to solve. Following spectral hashing \cite{Spectral-hash}, we relax the constraints $y_i \in \{-1,1\}^P$ to be $y_i=W^T \hat{K}_i + b$, then we have
\begin{displaymath}\label{eq:relax-obj}
\begin{split}
&\min_{W,\hat{K},b,E^{(m)}} \sum_{m=1}^M \mathcal{L}(Y,Y^{(m)}) +\alpha ||\hat{K}||_\ast + \lambda \sum_{m=1}^M ||E^{(m)}||_{2,1}, \\
&s.t.~~ K^{(m)}=\hat{K} + E^{(m)}, \hat{K} \geq 0, m=1,\ldots, M; \\
&y_i=W^T \hat{K}_i + b, YY^T=I.
\end{split}
\end{displaymath}

We rewrite the objective function by further minimizing the least square error regarding $Y$ while regularizing $W$ coupled with trade-off parameters $\beta$ and $\delta$, it then has
\begin{equation}\label{eq:rewrite-obj}
\begin{split}
&\min_{W,\hat{K},b,E^{(m)}} \sum_{m=1}^M \mathcal{L}(Y,Y^{(m)}) +\alpha ||\hat{K}||_\ast + \lambda \sum_{m=1}^M ||E^{(m)}||_{2,1} \\
&+ \beta \left( ||\hat{K}^T W+ \textbf{1}b-Y||_F^2 +\delta ||W||_F^2 \right)\\
&s.t.~~ K^{(m)}=\hat{K} + E^{(m)}, \hat{K} \geq 0, m=1,\ldots,M;  YY^T=I.
\end{split}
\end{equation}

Eq.\eqref{eq:rewrite-obj} is still non-convex due to orthogonal constraint $YY^T=I$.
Fortunately with either $W$, $b$ or $\hat{K}$, $E^{(m)}$ fixed, the problem is convex with respect to the other variables. Therefore, we present an alternating optimization way that can efficiently find the optimum in a few steps. First, given $\hat{K}$ and $E^{(m)}$, we show that computation expressions of $W$ and $b$ can be obtained.
To compute $\hat{K}$ and $E^{(m)}$, we employ an efficient optimization technique, the inexact augmented Lagrange multiplier (ALM) algorithm \cite{ALM-report}.

\section{Optimization}\label{sec:opt}

\subsection{Compute $W$ and $b$}
With other variables fixed, and setting the derivative of Eq.\eqref{eq:rewrite-obj} w.r.t. $b$ to zero, we get
\begin{equation}\label{eq:compute-b}
\begin{split}
&\textbf{1}^T \left(\hat{K}^T W+ \textbf{1}b-Y\right)=0 \\
&\Rightarrow b=\frac{1}{N}\left(\textbf{1}^T Y-\textbf{1}^T \hat{K}^T W\right).
\end{split}
\end{equation}
Setting the derivative of Eq.\eqref{eq:rewrite-obj} w.r.t. $W$ to zero, we yield
\begin{equation}\label{eq:compute-W}
\hat{K}(\hat{K}^T W+ \textbf{1}b-Y) +\delta W=0.
\end{equation}
Substituting $b$ in Eq.\eqref{eq:compute-b} into Eq.\eqref{eq:compute-W}, we have
\begin{equation}\label{eq:thiswb}
\begin{split}
&\hat{K}\hat{K}^T W+\hat{K} \textbf{1}\left(\frac{1}{N}(\textbf{1}^T Y-\textbf{1}^T \hat{K}^T W)\right)-\hat{K}Y=0\\
& \Rightarrow W=(\hat{K}L_c \hat{K}^T+\delta I)^{-1}+\hat{K}L_cY,
\end{split}
\end{equation}
where $L_c=I-\frac{1}{N}\textbf{11}^T$ is the centering matrix, and $L_c=L_c^T=L_cL_c^T$.

\subsection{Compute $\hat{K}$ and $E^{(m)}$}

With variables $W$ and $b$ being fixed, the problem turns to be
\begin{equation}\label{eq:solve-K-E}
\begin{split}
&\min_{\hat{K},E^{(m)}} \alpha ||\hat{K}||_\ast + \lambda \sum_{m=1}^M ||E^{(m)}||_{2,1}, \\
&s.t.~~ K^{(m)}=\hat{K} + E^{(m)}, \hat{K} \geq 0, m=1,\ldots,M.
\end{split}
\end{equation}
The rank minimization problem has been well studied in literature \cite{Seg-LRR,Robust-PCA}.  By introducing an auxiliary variable $Q$ such that $\hat{K} = Q$,  Eq.\eqref{eq:solve-K-E} can be then converted into the following equivalent form:
\begin{equation}
\begin{split}
&\mathcal{D}(\hat{K},Q, E^{(m)})=\alpha ||Q||_\ast +\lambda \sum_{m=1}^M ||E^{(m)}||_{2,1} \\
&+ \sum_{m=1}^M \left( \langle A^{(m)}, \hat{K} + E^{(m)}-K^{(m)} \rangle + \frac{\mu}{2} ||\hat{K} + E^{(m)}-K^{(m)}||_F^2 \right)\\
&+\langle B, \hat{K}-Q\rangle + \frac{\mu}{2}||\hat{K}-Q||_F^2,
\end{split}
\end{equation}
where $A^{(m)}$ and $B$ represent the Lagrange multipliers, $\langle \cdot,\cdot\rangle$ denotes the inner product of matrices, and $\mu>0$ is an adaptive penalty parameter.
Next we will elaborate the update rules for each of $\hat{K}$, $Q$, and $E^{(m)}$ by minimizing $\mathcal{D}$ while fixing the others.

\paragraph{Solving for $Q$}
When the other variables are fixed, the subproblem w.r.t. $Q$ is
\begin{equation}
\min_{Q} ||Q||_\ast + \frac{\mu}{2\alpha}||\hat{K}-Q +\frac{B}{\mu\alpha}||_F^2.
\end{equation}
It can be solved by the Singular Value Threshold method \cite{Singular-value}. More specifically, let $U\Sigma V^T$ be the SVD form of $(\hat{K}+\frac{B}{\mu\alpha})$, the updating rule of $Q$ using the SVD operator in each iteration will be
\begin{equation}\label{eq:update-Q}
Q=U\mathcal{S}_{1/\mu\alpha}(\Sigma)V^T,
\end{equation}
where $\mathcal{S}_{\varrho}(x)=\max(x-\varrho,0)+\min(x+\varrho,0)$ is the shrinkage operator \cite{ADMM}.

\paragraph{Solving for $E^{(m)}$}
The subproblem with respect to $E^{(m)}, (m=1,\ldots,M)$ can be simplified as
\begin{equation}\label{eq:update-E}
\min_{E^{(m)}} \lambda ||E^{(m)}||_{2,1} + \frac{\mu}{2} ||E^{(m)}-(K^{(m)}-\hat{K}-\frac{A^{(m)}}{\mu})||_F^2,
\end{equation}
which enjoys a closed form solution $E^{(m)}=\mathcal{S}_{\lambda/\mu}(K^{(m)}-\hat{K}-\frac{A^{(m)}}{\mu})$.

\paragraph{Solving for $\hat{K}$}
With the other variables being fixed, we update $\hat{K}$ by solving
\begin{equation}\label{eq:update-K}
\hat{K}=\arg\min_{\hat{K}} ||\hat{K}+E^{(m)}-K^{(m)}+\frac{A^{(m)}}{\mu}||_F^2 + \frac{\mu}{2}||\hat{K}-Q+\frac{B}{\mu}||_F^2
\end{equation}
For ease of representation, we define $C=\frac{1}{M}(Q-\frac{B}{\mu}+\sum_{m=1}^M (K^{(m)}-E^{(m)}-\frac{A^{(m)}}{\mu}))$. Then, the problem in Eq.\eqref{eq:update-K} can be rewritten as
\begin{equation}\label{eq:sub-K}
\begin{split}
&\hat{K}=\arg\min_{\hat{K}}\frac{1}{2}||\hat{K}-C||_F^2=\arg\min_{\hat{K}_1,\ldots,\hat{K}_N}=\frac{1}{2}\sum_{i=1}^N||\hat{K}_i-C_i||_2^2.\\
& s.t. \hat{K} \geq 0, \hat{K}_{i (i = 1,\ldots, N)} \geq 0.
\end{split}
\end{equation}

Hence, the problem in Eq.\eqref{eq:sub-K} can be decomposed into $N$ independent subproblems: $\min_{\hat{K}_i}\frac{1}{2}||\hat{K}_i-C_i||_2^2$, subject to $\hat{K}_i \geq 0$.   Each subproblem is a proximal operator problem, which can be efficiently solved by the projection algorithm in \cite{Duchi-ICML-2008}.

\subsection{Learning Hash Codes}
Once the hashing function implemented by $W$ and $b$ is learned by exploiting the kernelized similarity consensus $\hat{K}$, we can generate hash codes for both database and query samples, denoted as $x_t$, via Eq.~\eqref{eq:hash-code}.
\begin{equation}\label{eq:hash-code}
y_t = sign(W^T [\mathcal{K}(x_t,Z_1),\ldots, \mathcal{K}(x_t, Z_R)]^T + b),
\end{equation}
where $W \in \mathbb{R}^{R  \times P}$, $\mathcal{K}(x_t,Z_i)$ represents the similarity between $x_t$ and the $i$-th landmark using Gaussian RBF kernel over the concatenated feature space for all views.


\subsection{Out-of-Sample Extension}

An essential part of hashing is to generate binary codes for new samples, which is known as out-of-sample problems. A widely used solution is the Nystr$\ddot{o}$m extension \cite{Nystrom}. However, this is impractical for large-scale hashing since the Nystr$\ddot{o}$m extension is as expensive as doing exhaustive nearest neighbor search with a complexity of $\mathcal{O}(N)$ for $N$ data points. In order to address the out-of-sample extension problem, we employ a non-parametric regression approach, inspired by Shen \etal \cite{InductiveHash}. Specifically, given the hashing embedding $Y = \{y_1,y_1,\dots, y_N\}$ for the entire training set $X=\{x_1,x_2,\dots,x_N\}$, for a new data point $x_q$, we aim to generate a hashing embedding $y_q$ while preserving the local neighborhood relationships among its neighbors $\mathcal{N}_k(x_q)$ in $X$. A simple inductive formulation can produce the embedding for a new data point by a sparse linear combination of the base embeddings:

\begin{equation}\label{eq:inductive}
y_q=\frac{\sum_{i=1}^N w(x_q,x_i) y_i}{\sum^N_{i=1}w(x_q,x_i)},
\end{equation}
where we define \[w(x_1,x_i)=\left\{\begin{array}{cl}
exp(-||x_q - x_i||^2/\sigma^2), & x_i \in \mathcal{N}_k(x_q)\\
0,& \mbox{elsewhere.}\end{array}\right.\]

However, Eq.\eqref{eq:inductive} does not scale well for computing out-of-sample extension ($\mathcal{O}(N)$) for large-scale tasks. To this end, we employ a prototype algorithm \cite{InductiveHash} to approximate $y_q$ using only a small base set:

\begin{equation}\label{eq:approximate}
h(x_q) = sign\left(\frac{\sum_{j=1}^Z w(x_q,c_j) y_j}{\sum^Z_{j=1}w(x_q,c_j)}\right)
\end{equation}
where $sign(\cdot)$ is the sign function, and $Y=\{y_1,y_2,\dots,y_Z\}$ is the hashing embedding for the base set $B=\{c_1,c_2,\dots,c_Z\}$ which is the cluster centers obtained by K-means. In this stage, the major computation cost comes from K-means clustering, which is $\mathcal{O}(dlZN)$ in time ($d$ is the feature dimension, and $l$ is the number of iterations in K-means). The iteration number $l$ can be set less than 50, thus, the K-means only costs $\mathcal{O}(dZN)$. Considering that $Z$ is much less than $N$, the total time is linear in the size of training set. The computation of distance between $B$ and $X$ cost $\mathcal{O}(dZN)$. Thus, the overall time cost is $\mathcal{O}(dZN + dZN) = \mathcal{O}(dZN)$.

\section{Complexity Analysis}\label{sec:complexity}

We analyze the time complexity regarding per iteration of the optimization strategy.
The complexity of computing $\hat{K}L_c \hat{K}^T$ and $(\hat{K}L_c \hat{K}^T+\delta I)^{-1}$ in Eq.\eqref{eq:thiswb} is $\mathcal{O}(R^2 N)$ and $\mathcal{O}(R^3)$, respectively. Commonly, $R (\ll N)$ landmarks are generated off-line via scalable K-means clustering for less than 50 iterations, keeping the complexity of computing $W$ to be $\mathcal{O}(R^2 N)+\mathcal{O}(R^3)$.
The complexity of computing hash codes for a new sample is $\mathcal{O}(dZN)$.  Overall, the time complexity is $\mathcal{O}(dZN+R^3+R^2N) \approx \mathcal{O}(dZN + R^2N)$ in one iteration, which is linear with respect to the training size.

\section{Experiments}\label{sec:exp}

\subsection{Experimental Settings}

\paragraph{Competitors} We compare our method with recently proposed state-of-the-art multiple feature hashing algorithms:
\begin{itemize}
\item Multiple feature hashing (\textbf{MFH}) \cite{MM11}: This method exploits local structure in each feature and global consistency in the optimization of hashing functions.
\item Composite hashing with multiple sources (\textbf{CHMS}) \cite{Composite-hash}: This method treats a linear combination of view-specific similarities as an average similarity which can be plugged into a spectral hashing framework.
\item Compact kernel hashing with multiple features (\textbf{CKH}) \cite{MM12}: It is a multiple feature hashing framework where multiple kernels are linearly combined.
\item Sequential spectral hashing with multiple representations (\textbf{SSH}) \cite{ECCV12}: This method constructs an average similarity matrix to assemble view-specific similarity matrices.
\item Multi-View Latent Hashing (\textbf{MVLH}) \cite{MVLH}: This is an unsupervised multi-view hashing approach where binary codes are learned by the latent factors shared by multiple views from an unified kernel feature space.
\end{itemize}

\paragraph{Datasets} We conduct the experiments on two image benchmarks: CIFAR-10 \footnote{http://www.cs.toronto.edu/~kriz/cifar.html} and NUS-WIDE.
\begin{itemize}
\item \textbf{\underline{CIFAR-10}} consists of 60K 32$\times$32 color images from ten object categories, each of which contains 6K samples. Every image is assigned to a mutually exclusive class label and for each image, we extract 512-dimensional GIST feature \cite{GIST} and 300-dimensional bag-of-words quantized from dense SIFT features \cite{SIFT} to be \emph{two views}.
\item \textbf{\underline{NUS-WIDE}} \cite{NUS-WIDE} contains 269,648 labeled images crawled from Flickr and is manually annotated with 81 categories.  Three types of features are extracted: 128-dimensional wavelet texture, 225-dimensional block-wise color moments, and 500-dimensional bag-of-words to construct \emph{three views}.
\end{itemize}

\paragraph{Multi-view Corruption Setting} In \textbf{CIFAR-10}, considering that missing features may have some structure, we remove a square patch of pixels from each image covering 25\% of the total number of pixels. The location of the patch is uniformly sampled for each image. This will naturally deteriorate view-dependent feature representations. In \textbf{NUS-WIDE}, we consider the scenario where 20\% of feature values in each view are corrupted with perturbation noise following a standard Gaussian distribution.

\paragraph{Parameter Setting}
In the training phase, we uniformly sample 30K and 100K images as training data from both datasets, and generate 300 and 500 landmarks. That is, we fix the graph construction parameters $L=300$, $k=3$ on CIFAR-10, and $L=500$, $k=5$ on NUS-WIDE, respectively. In the testing phase, we randomly select 1,000 query images in which the true neighbors of each image are defined as the semantic neighbors which share at least one common semantic label. For our method and \textbf{CKH}, we use Gaussian RBF kernel $K^{(i)}(x,y)=\exp(-||x-y||_i^2/2\sigma^2)(i = 1,\cdot,M)$, where $||\cdot||_i^2$ represents the Euclidean distance within the $i$-th feature space. The parameter $\sigma$ is learned via the self-tuning strategy \cite{STSC}.

In Eq.\eqref{eq:rewrite-obj}, there are five tunable parameters: $\gamma$, $\delta$, $\alpha$, $\beta$, and $\lambda$.
Parameters $\gamma$ and $\delta$ controlling global hash code learning and regularization on hashing functions are set as $10^{-4}$ and $10^{-6}$, respectively.
For $\alpha$, $\beta$, and $\lambda$, we tune their optimal combination, that is, $\alpha=10^{-1}$, $\beta=10^{0}$, and $\lambda=10^{-3}$, as conducted in section \ref{ssec:parameter}.

\paragraph{Evaluation Metric}  The mean precision-recall and mean average precision (MAP) are computed over the retrieved set consisting of the samples with the hamming distance \cite{Supervised-hash-kernel} using 8 to 32 bits to a specific query.
We carry out hash lookup within a Hamming radius 2 and report the mean hash lookup precision over all queries.
For a query $q$, the average precision (AP) is defined as $AP(q)=\frac{1}{L_q}\sum_{z=1}^l P_q(z)\varpi_q(z)$, where $L_q$ is the number of ground-truth neighbors of $q$ in database, $l$ is the number of entities in database, $P_q(z)$ denotes the precision of the top $z$ retrieved entities, and $\varpi_q(z)=1$ if the $z$-th retrieved entity is a ground-truth neighbor and $\varpi_q(z)=0$, otherwise. Ground truth neighbors are defined as items which share at least one semantic label. Given a query set of size $F$, the MAP is defined as the mean of the average precision for all queries: $MAP=\frac{1}{F}\sum_{i=1}^F AP(q_i)$.

\subsection{Results}

\begin{table*}[t]\scriptsize
\caption{Hash lookup precision (mean$\pm$std) with Hamming radius 2 on different databases.}\label{tab:lookup}
\hspace{-2cm}
\begin{center}
\begin{tabular}{|c|c|c|c|c|c|c|c|c|}
\hline
\hline
\multirow{2}{*}{Method} & \multicolumn{4}{|c|}{CIFAR-10} & \multicolumn{4}{|c|}{NUS-WIDE}\\
\cline{2-9}
& P=8 & P=32 & P=48 & P=128 & P=8 & P=32 & P=48 & P=128\\
\hline
\hline
\textbf{ MFH} & 23.31$\pm$0.71 & 28.19$\pm$0.48 & 26.38$\pm$0.68 & 23.68$\pm$0.71 & 23.52$\pm$0.72 & 26.49$\pm$0.85 & 33.55$\pm$0.49 & 34.97$\pm$0.81 \\
\cline{2-9}
\textbf{ CHMS} & 25.61$\pm$0.22 & 31.8$\pm$0.66 & 26.54$\pm$0.52 & 19.38$\pm$0.84 & 27.54$\pm$0.41 & 30.22$\pm$0.92 & 28.24$\pm$0.96 & 27.52$\pm$1.12 \\
\cline{2-9}
\textbf{ CKH} & 31.75$\pm$0.53 & 32.05$\pm$0.72 & 37.32$\pm$0.76 & 34.45$\pm$0.81 & 29.72$\pm$0.43 & 37.84$\pm$0.63 & 33.56$\pm$0.82 & 34.42$\pm$1.32 \\
\cline{2-9}
\textbf{ SSH} & 27.34$\pm$0.46 & 35.78$\pm$0.68 & 29.36$\pm$0.63 & 27.52$\pm$0.72 & 28.95$\pm$0.46 & 33.42$\pm$0.88 & 30.05$\pm$0.71 & 29.21$\pm$0.98 \\
\cline{2-9}
\textbf{ MVLH} & 32.27$\pm$0.41 & 40.24$\pm$0.63 & 44.81$\pm$0.46 & 42.06$\pm$0.62 & 31.92$\pm$0.62 & 39.05$\pm$0.87 & 40.31$\pm$0.52 & 36.12$\pm$0.70 \\
\cline{2-9}
\textbf{ Ours} & \textbf{36.73$\pm$0.41} & \textbf{47.63$\pm$0.52} & \textbf{51.22$\pm$0.36} & \textbf{46.57$\pm$0.44} & \textbf{34.21$\pm$0.48} & \textbf{46.35$\pm$0.47} & \textbf{44.33$\pm$0.34} & \textbf{43.08$\pm$0.32}\\
\hline
\end{tabular}
\end{center}
\end{table*}

\begin{figure}[t]
\begin{tabular}{cc}
\includegraphics[width=3.5cm]{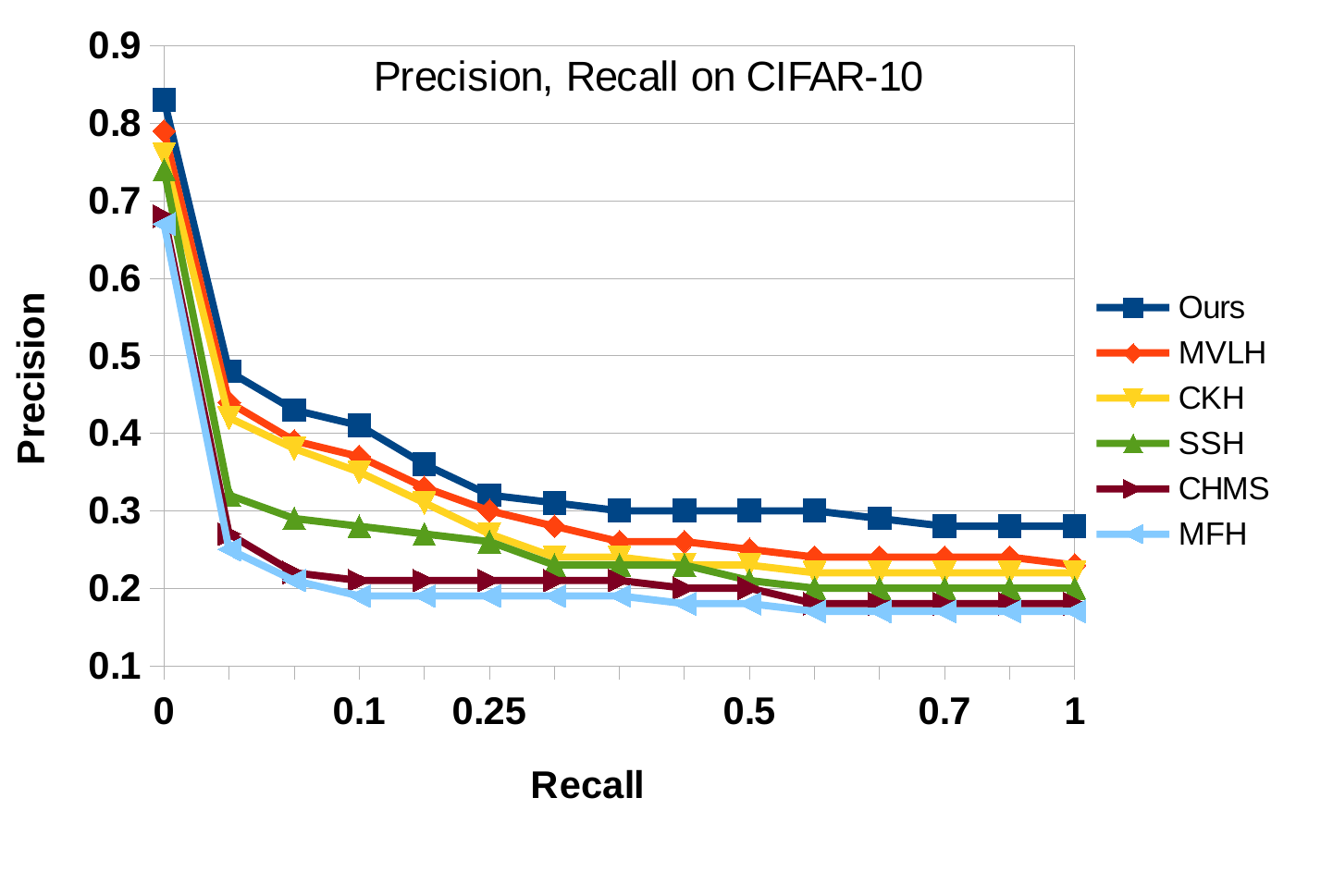}&
\includegraphics[width=3.5cm]{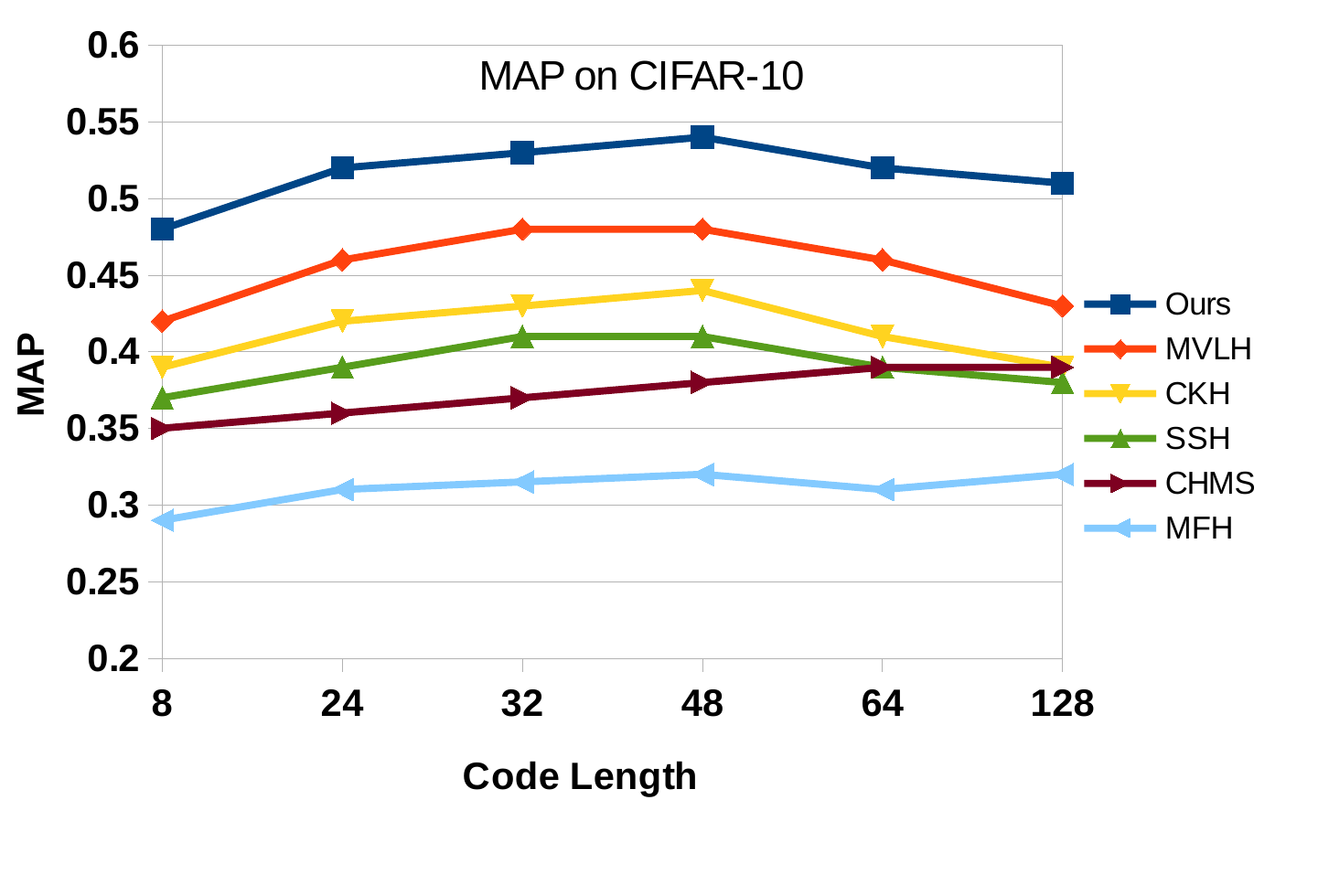}\\
\end{tabular}
\caption{Performance comparison on CIFAR-10 database. Left: Mean precision-recall of Hamming ranking at 48 bits. Right: Mean average precision of Hamming ranking w.r.t. 8-128 bits.}\label{fig:results-CIFAR-10}
\end{figure}

\begin{figure}[t]
\begin{tabular}{cc}
\includegraphics[width=3.5cm]{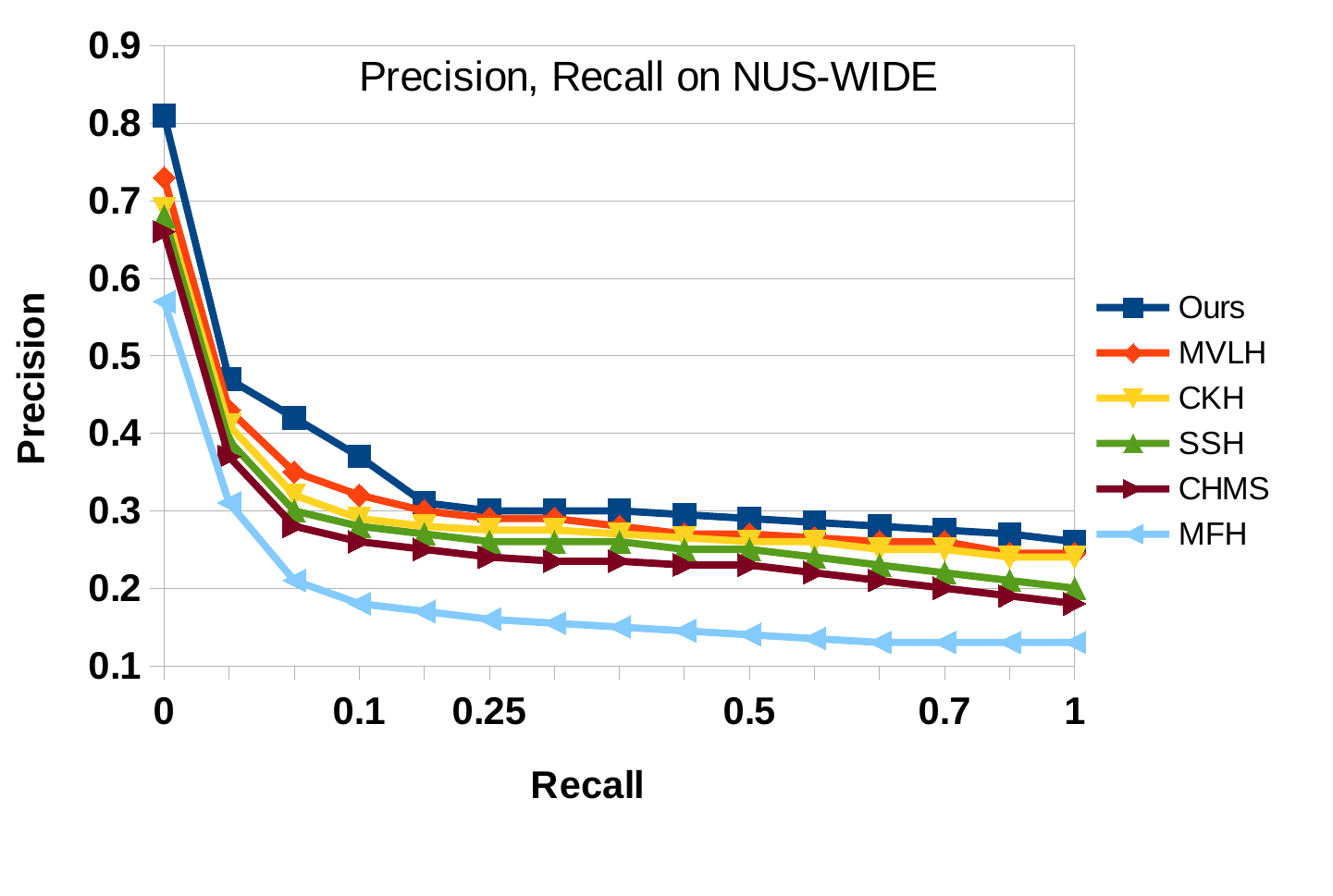}&
\includegraphics[width=3.5cm]{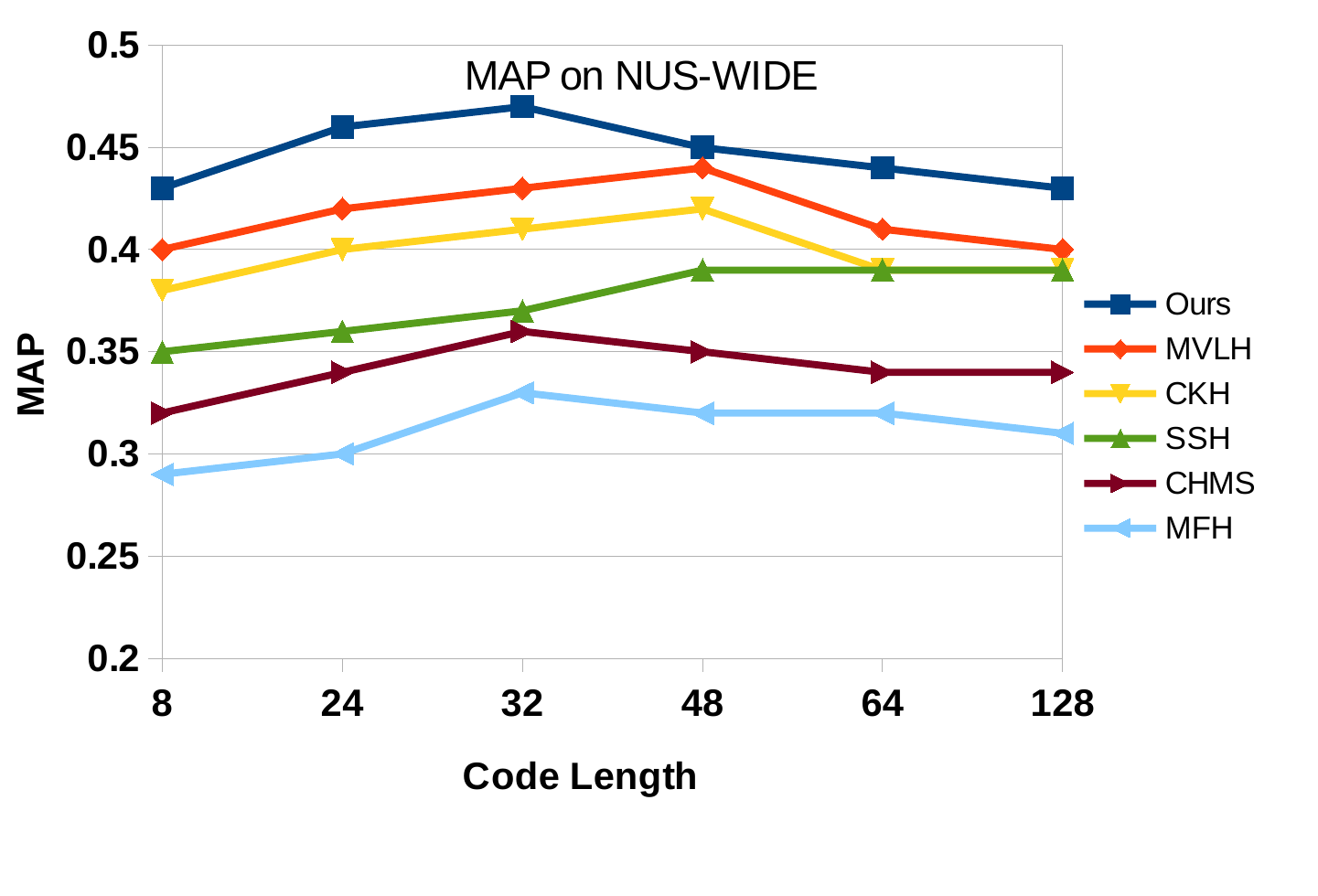}\\
\end{tabular}
\caption{Performance comparison on NUS-WIDE database. Left: Mean precision-recall of Hamming ranking at 64 bits. Right: Mean average precision of Hamming ranking w.r.t. 8-128 bits.}\label{fig:results-NUS-WIDE}
\end{figure}

\begin{figure*}[t]
\centering
\begin{tabular}{ccc}
\includegraphics[width=4.5cm]{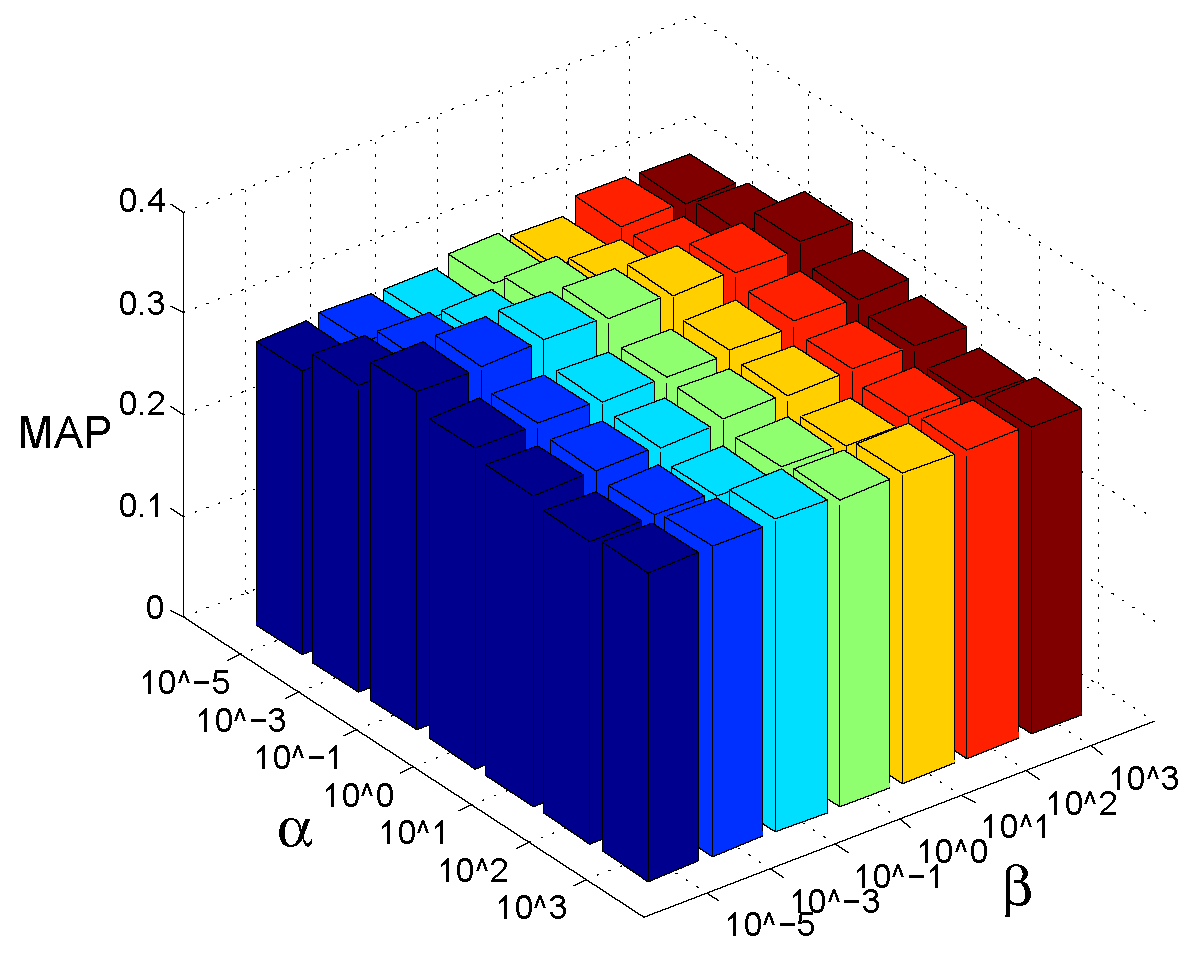}&
\includegraphics[width=4.5cm]{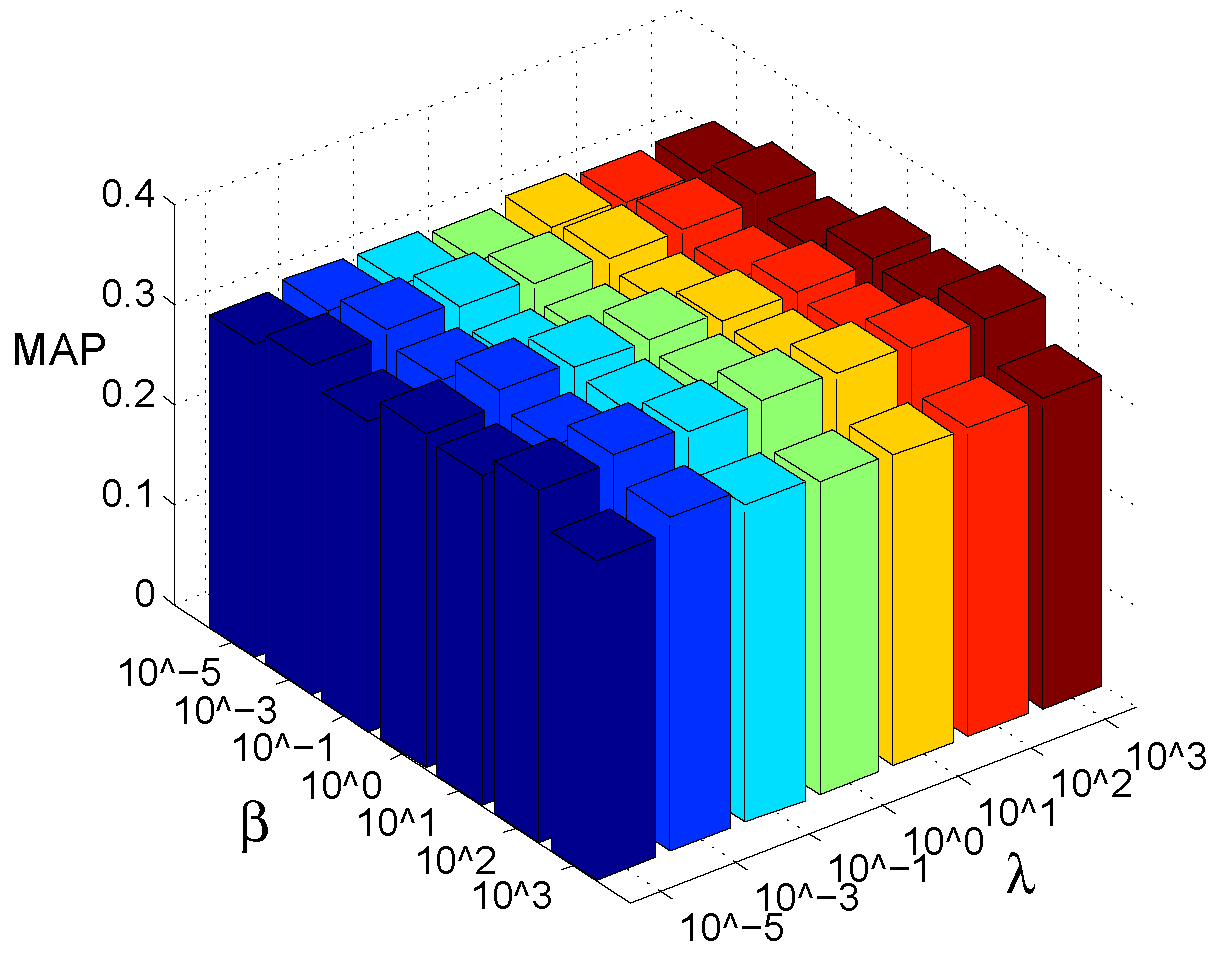}&
\includegraphics[width=4.5cm]{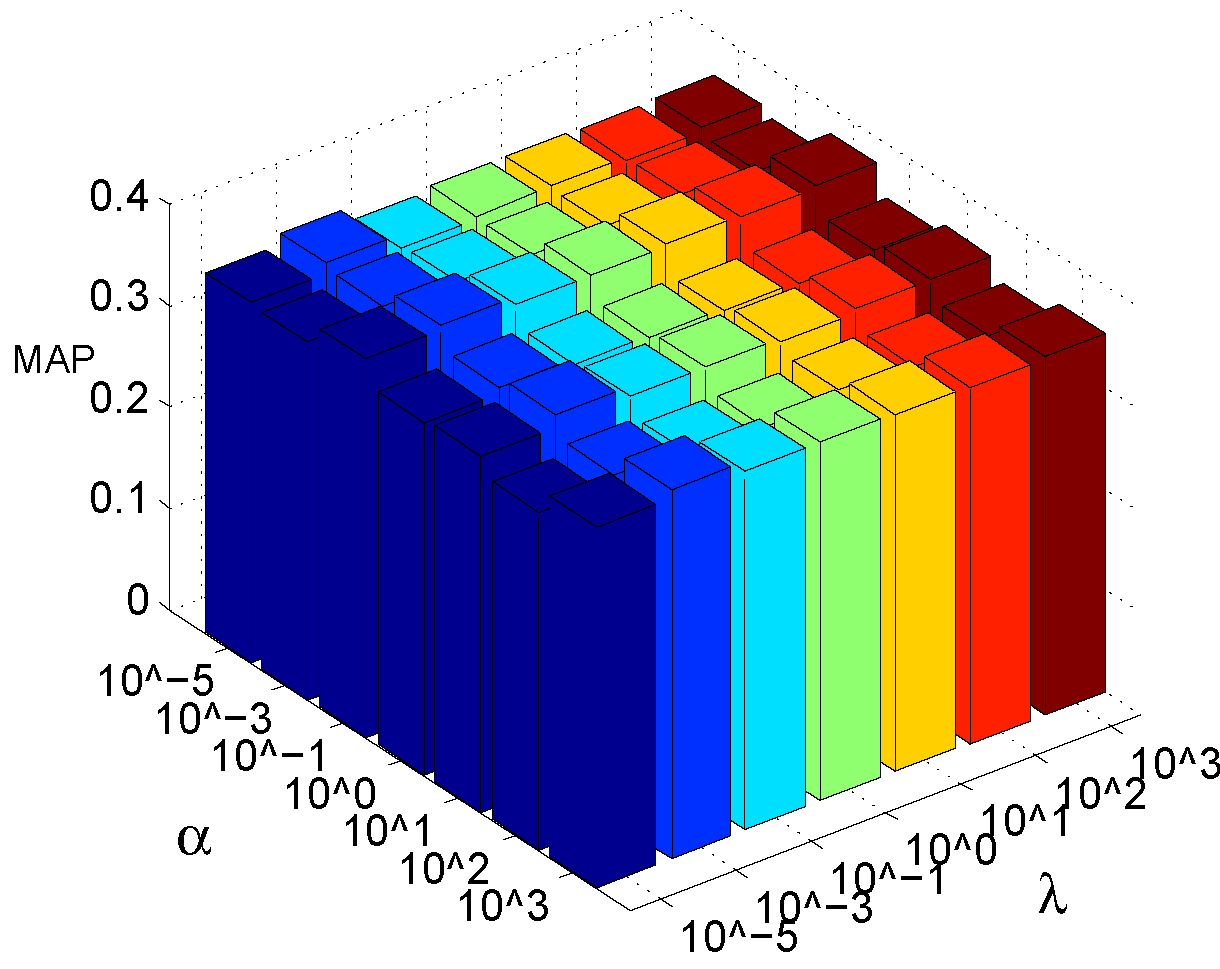}\\
(a) $\lambda$ is fixed as $10^{-3}$ & (b) $\alpha$ is fixed as $10^{-1}$ & (c) $\beta$ is fixed as $10^0$
\end{tabular}
\caption{The MAP variations of different parameter settings on NUS-WIDE database.}\label{fig:parameter}
\end{figure*}

We report the mean precision-recall curves of Hamming ranking, and mean average precision (MAP) w.r.t. different number of hashing bits over 1K query images. Results are shown in Fig.\ref{fig:results-CIFAR-10}, which are computed from top-100 retrieved samples. It can be seen from top subfigure of Fig.\ref{fig:results-CIFAR-10} that our method achieves a performance gain in both precision and recall over all counterparts and the second best is \textbf{MVLH}. This can demonstrate the superiority of using nonlinear hashing functions in nonlinear space. More importantly, the latent consensus kernelized similarity matrix by low-rank minimization is not only effective in leveraging complementary information from multi-views, but also robust against the presence of errors.
The subfigure (bottom) in Fig.\ref{fig:results-CIFAR-10} shows that as the hashing bit number varies, our method consistently keeps superior performance. Specifically, it reaches the highest precision value for 48 bits and shows a relatively steady performance with more hashing bits.
The results from the NUS-WIDE database are shown in Fig.\ref{fig:results-NUS-WIDE}. Once again we can see performance gaps in precision-recall between our approach and competitors, as illustrated in top subfigure of Fig.\ref{fig:results-NUS-WIDE}. This validates the advantage of our method by exploiting consensus of kernelized similarity to learn robust nonlinear hashing functions. In subfigure (bottom) of Fig.\ref{fig:results-NUS-WIDE}, as the number of hashing bit increases, our method is able to keep high and steady MAP values.

\begin{table}\scriptsize
\centering
\caption{Training/test time comparison on different algorithms using 64 bits. All training/test time is recorded in second. The training size of two datasets are 30K and 100K, respectively.}\label{tab:time}
\begin{tabular}{|c|c|c|c|c|}
\hline
\hline
\multirow{2}{*}{Method} & \multicolumn{2}{|c|}{CIFAR-10} & \multicolumn{2}{|c|}{NUS-WIDE}\\
\cline{2-5}
& Training  & Test & Training & Test\\
\hline
\textbf{MFH} & 32.8 & 6.4$\times 10^{-5}$ & 41.6 & 8.5$\times 10^{-5}$\\
\cline{2-5}
\textbf{CHMS} & 29.8 & 4.7$\times 10^{-5}$ & 37.2 & 7.8$\times 10^{-5}$\\
\cline{2-5}
\textbf{SSH} & 23.6 & 1.3$\times 10^{-5}$ & 31.7 & 2.4$\times 10^{-5}$\\
\cline{2-5}
\textbf{CKH} & 10.7 & 2.3$\times10^{-6}$ & 15.3 & 3.2$\times 10^{-6}$\\
\cline{2-5}
\textbf{MVLH} & 20.4 & 2.2$\times10^{-6}$ & 28.1 & 4.3$\times 10^{-6}$\\
\cline{2-5}
\textbf{Ours} & 14.1 & 2.6$\times10^{-6}$ & 19.2 & 3.5$\times 10^{-6}$\\
\cline{2-5}
\hline
\end{tabular}
\end{table}

To evaluate the impact of hashing bit numbers on performance of hash lookup, in Table \ref{tab:lookup}, we report hash lookup mean precision with standard deviation (mean$\pm$std) in the case of 8, 32, 48, 128 bits on both databases. Similar to Hamming ranking results, our method achieves the better performance than others and obviously increasing performance with less than 32 bits, which demonstrates that our approach with compact hashing codes can retrieve more semantically related images than all baselines in terms of hash lookup.

In Table \ref{tab:time}, we report the comparison on training/test time over the two image benchmarks. \textbf{CKH} and our method are much more efficient by taking less than 15s and 20s respectively to train on CIFAR-10 and NUS-WIDE using 32 bits. The efficiency improvement comes from the usage of landmarks. While our method is slightly less efficient to \textbf{CKH} because of the low-rank kernelized similarity recovery, it is very comparable to \textbf{CKH} and consistently superior to \textbf{CKH} in other performance. \textbf{MVLH} is relatively costly due to its expensive matrix factorization in its kernel space. \textbf{MFH} and \textbf{CHMS} are time-consuming in training stage because they both involve the eigen-decomposition of a dense affinity matrix, which is not scalable to a large-scale setting. \textbf{SSH} has a gain in efficiency compared with \textbf{MFH} and \textbf{CHMS} on account of their approximation on the K-nearest graph construction \cite{ECCV12}.

\subsection{Parameter Tuning}\label{ssec:parameter}

\begin{figure}
\centering
\begin{tabular}{c}
\includegraphics[width=6cm]{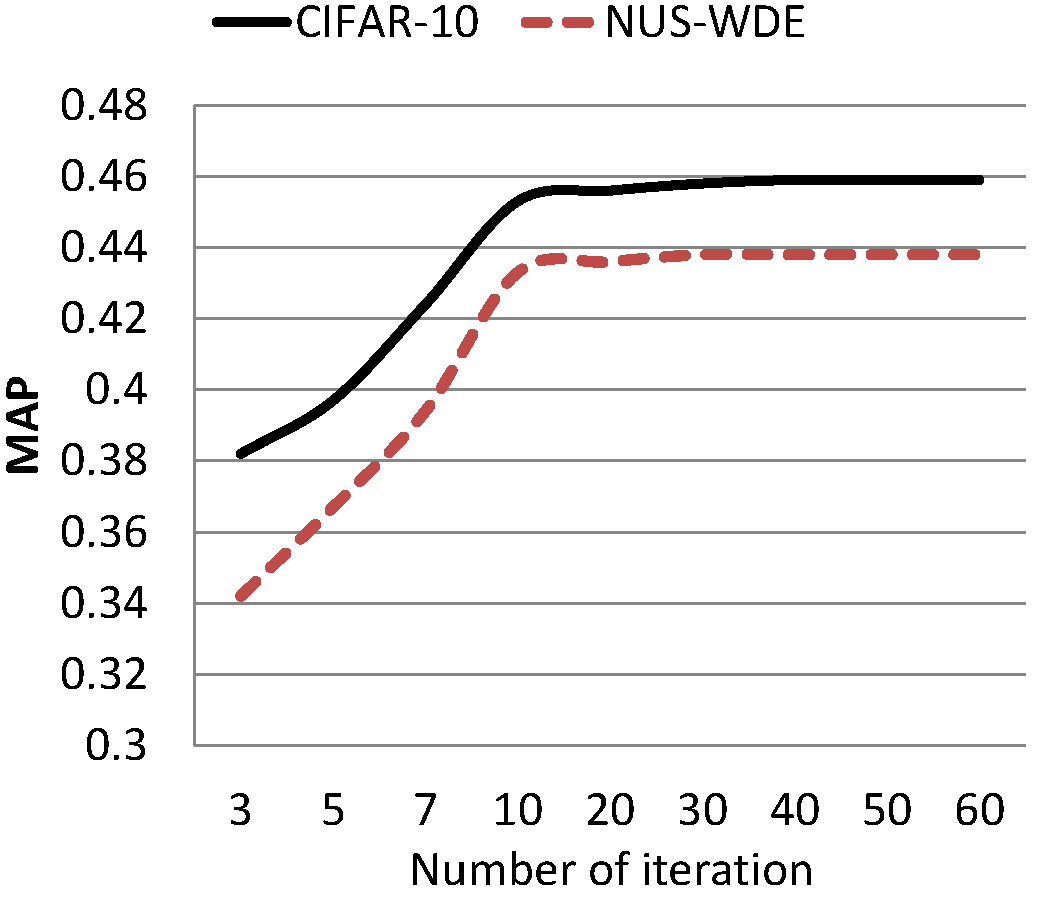}\\
\end{tabular}
\caption{Convergence study over real-world datasets.}\label{fig:robust-convergence}
\end{figure}

In this experiment, we test different parameter settings for our algorithm to study the performance sensitivity. We learn three parameters: $\alpha$, $\lambda$, and $\beta$, corresponding to the term of requisite component, non-requisite decomposition, and hashing function learning in Eq.\eqref{eq:rewrite-obj}. For these parameters, we tune them from $\{10^{-5},10^{-3},10^{-1},10^0,10^1,10^2,10^3\}$. We fix one of the parameters in $\alpha$, $\lambda$, and $\beta$ to report the MAP while the other two parameters are changing. The results are shown in Fig.\ref{fig:parameter}. In Fig.\ref{fig:parameter} (a), by fixing $\lambda=10^{-3}$, we show the performance variance on different pairs of $\alpha$ and $\beta$. We can observe that our algorithms achieves a relatively higher MAP when $\alpha=0.1$, and $\beta=10^{0}$. The similar performance can also be seen from Fig.\ref{fig:parameter} (b) and Fig.\ref{fig:parameter} (c). Thus, among different combinations, the method gains the best performance when $\alpha=10^{-1}$, $\beta=1$, and $\lambda=10^{-3}$, while it is relatively insensitive to varied parameters setting. With optimal combination of parameters, we study the issue of convergence. In Fig.\ref{fig:robust-convergence}, we can observe that our algorithm becomes convergent in less than 40 iterations, demonstrating its fast convergence rate.

\subsection{Out-of-Sample Case}\label{ssec:out-of-sample}

\begin{figure}
\begin{tabular}{cc}
\includegraphics[width=3.5cm]{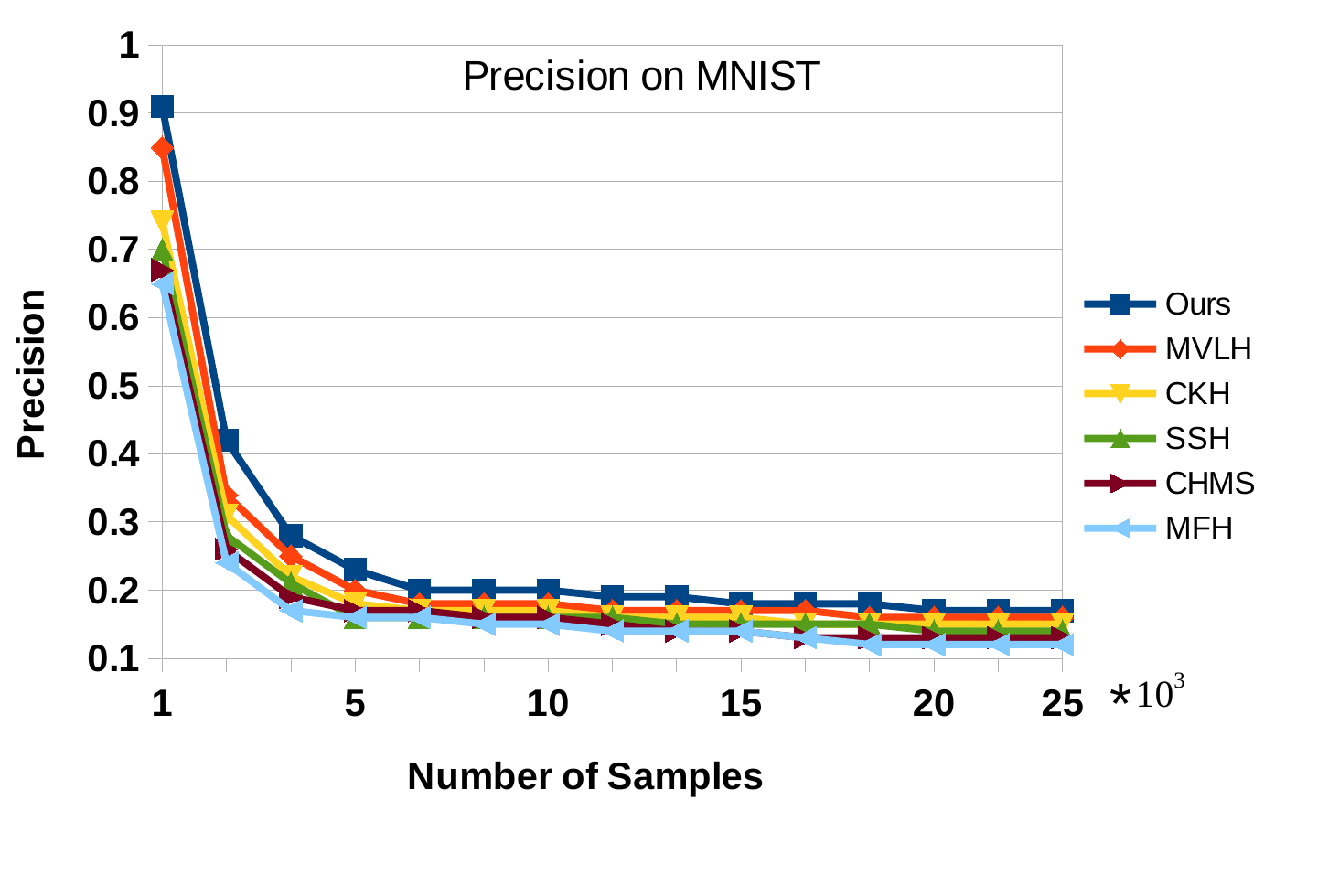}&
\includegraphics[width=3.5cm]{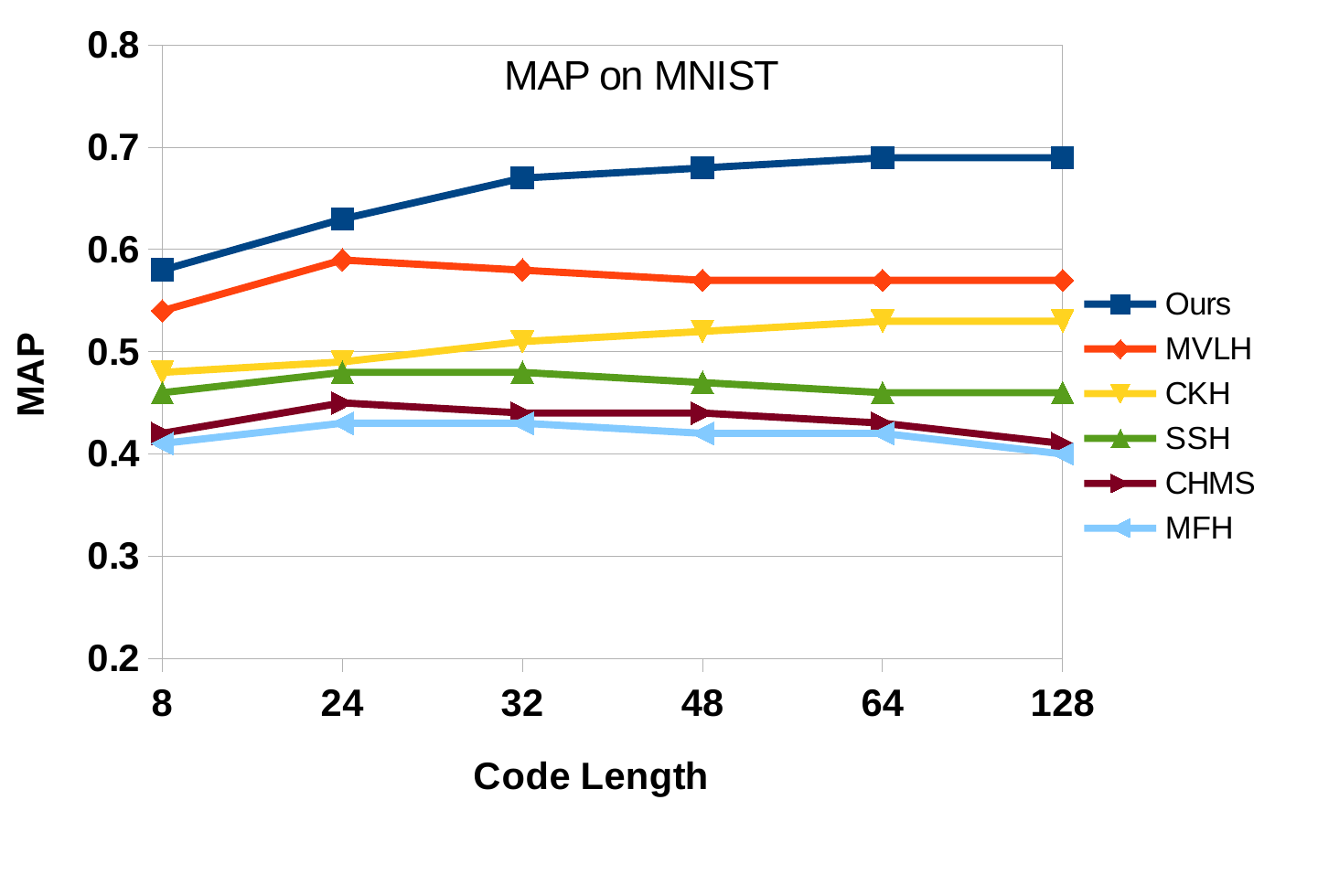}\\
\end{tabular}
\caption{Performance comparison on MNIST database using base hash functions learn from CIFAR-10 dataset. Left: Precision with respect to number of returned sampled at 64 bits. Right: Mean average precision of Hamming ranking w.r.t. 8-128 bits.}\label{fig:results-MNIST}
\end{figure}

In this experiment, we study the property of out-of-sample extension. We take the CIFAR-10 dataset as the base benchmark to train base embeddings. Another dataset MNIST is considered as the testing bed. The MINIST dataset \cite{MNIST} consists of 70K images, each of 784 dimensions, of handwritten digits from ``0" to ``9". As in Fig.\ref{fig:results-MNIST}, our method achieves the best results. On this dataset, we can clearly see that our method outperforms \textbf{MVLH} by a large margin, which increases as code length increases. This further demonstrates the advantage of kernelized low-rank embedding as a tool for hashing by embedding high dimensional data into a lower dimensional space. This dimensionality reduction procedure not only preserves the local neighborhood, but also reveals global structure.

\section{Conclusion}\label{sec:con}
In this paper, we motivate the problem of robust hashing for similarity search over multi-view data objects under a practical scenario that error corruptions for view-dependent feature representations are presented. Unlike existing multi-view hashing methods that take a two-phase scheme of constructing similarity matrices and learning hash functions separately, we propose a novel technique to jointly learn hash functions and a latent, low-rank, corruption-free kernelized similarity under multiple representations with potential noise corruptions.
Extensive experiments conducted on real-world multi-view data sets demonstrate the superiority of our method in terms of efficacy.

\bibliographystyle{named}
\bibliography{ijcai16}

\end{document}